\pdfoutput=1
\documentclass[10pt,twocolumn,letterpaper]{article}

\usepackage{times}
\usepackage[dvipsnames]{xcolor}
\usepackage{graphicx}
\usepackage{amsmath,amssymb}
\usepackage{booktabs}
\usepackage[numbers,sort&compress]{natbib}
\usepackage{url}
\usepackage{enumitem}
\usepackage{currfile} %< for \currfilebase macro

\usepackage{caption} %< for "\captionof" commands (teaser)
\newcommand{\isolatedbest}[1]{\textbf{\textcolor{RoyalBlue}{#1}}}
\newcommand{\routedbest}[1]{\textbf{\textcolor{ForestGreen}{#1}}}
\DeclareRobustCommand{\tinybox}[2]{\begingroup\setlength{\fboxsep}{1.1pt}\colorbox{#1!12}{\textcolor{#1}{\scriptsize\bfseries #2}}\endgroup}
\DeclareRobustCommand{\dataicon}{\tinybox{RoyalBlue}{\ensuremath{\blacksquare}}}
\DeclareRobustCommand{\abcicon}{\tinybox{ForestGreen}{\ensuremath{\circlearrowright}}}
\DeclareRobustCommand{\attnicon}{\tinybox{Plum}{\ensuremath{\blacklozenge}}}
\DeclareRobustCommand{\routeicon}{\tinybox{BurntOrange}{\ensuremath{\blacktriangleright}}}
\DeclareRobustCommand{\ctrlcon}{\tinybox{RedOrange}{\ensuremath{\triangle}}}
\DeclareRobustCommand{\evalicon}{\tinybox{Gray}{\ensuremath{\bullet}}}
\DeclareRobustCommand{\bestcell}[1]{\textbf{\textcolor{ForestGreen}{#1}}}
\DeclareRobustCommand{\riskcell}[1]{\textbf{\textcolor{RedOrange}{#1}}}
\DeclareRobustCommand{\gaincell}[1]{\textbf{\textcolor{RoyalBlue}{#1}}}

\newenvironment{samepageblock}{\par\medskip\noindent\begin{minipage}{\linewidth}}{\end{minipage}\par\medskip}

\newcommand{\SamePageWideFigure}[4][.95\textwidth]{%
  \begin{figure*}[!t]
  \centering
  \begin{minipage}{#1}
  \centering
  \includegraphics[width=\linewidth]{#2}
  \caption{#3}
  \label{#4}
  \end{minipage}
  \end{figure*}}
\newenvironment{SamePageTable}[2]{%
  \begin{samepageblock}
  \centering
  \def\spcaption{#1}\def\splabel{#2}}{%
  \captionof{table}{\spcaption}
  \ifx\splabel\empty\else\label{\splabel}\fi
  \end{samepageblock}}

\usepackage[pagebackref,breaklinks,colorlinks,allcolors=black]{hyperref}
\usepackage[capitalize]{cleveref}

\title{Efficient Visual Pointing for Embodied AI:\\Agent-Driven Data Synthesis, Cross-Block Attention, and Iterative Correction}

\author{Zijian Hong\textsuperscript{1} \quad Qi Lv\textsuperscript{1} \quad Yuxiang Xie\textsuperscript{1} \quad Jianming Xing\textsuperscript{1}\\
Xiang Deng\textsuperscript{1} \quad Weili Guan\textsuperscript{1} \quad Liqiang Nie\textsuperscript{1}\\
\textsuperscript{1}Harbin Institute of Technology (Shenzhen)\\
{\tt\small \{25b952003@stu.hit.edu.cn, guanweili@hit.edu.cn\}}
}

\begin{document}
\maketitle

\begin{abstract}
Visual pointing maps a language instruction to pixel coordinates, a core skill for embodied AI. We describe our PointArena 2026 solution, which achieves 77.2\% overall accuracy and ranks second on the benchmark. The approach targets three failure modes. First, agent-driven synthesis builds large semantic and anchor-relative candidate pools; the server inventory contains 55,372 processed outputs, 53,772 de-duplicated sample IDs, and 37,574 trainable completed or accepted rows. Second, a deterministic steerable-data pipeline creates a verified 10,000-sample main set, plus reserve samples, using masks, templates, and path verification. Third, two model-side modules address complementary errors: AttnRes adds gated cross-block attention for steerability, while ABC correction encodes perturbed coordinates with visual features for general coordinate grounding. Category-aware routing combines complementary specialists; local validation used to select experts records 93.9\% Affordance, 82.6\% Spatial Relation, 78.2\% Reasoning, 70.4\% Counting, and 63.0\% Steerability.
\end{abstract}

\section{Introduction}
\label{sec:intro}

Visual pointing localizes natural-language instructions to image coordinates. This is stricter than image-text matching: the model must bind object identity, affordance, spatial relation, and sometimes multiple instances to a valid pixel. PointArena 2026~\cite{pointarena2026} stresses this capability with five categories: Affordance, Spatial Relation, Reasoning, Counting, and Steerability. The last category is especially different from common object-reference data because the target is defined relative to an anchor point and a direction.

Our starting model is a Molmo-style coordinate-generating VLM~\cite{molmo_pixmo}. Its zero-shot result is strong overall (72.7\%), but the category breakdown exposes three bottlenecks. \textbf{Data scarcity}: public coordinate data is dominated by object naming, while steerability has no direct public supervision. \textbf{Spatial state propagation}: standard LoRA fine-tuning~\cite{lora} does not explicitly pass an anchor state across transformer blocks. \textbf{One-shot prediction}: a single coordinate sample gives no structured way to correct a near miss.

We address these bottlenecks with a compact system summarized in \cref{fig:overview}. The contribution is not one monolithic model but a set of targeted interventions:
\begin{figure*}[t]
  \centering
  \includegraphics[width=.95\textwidth]{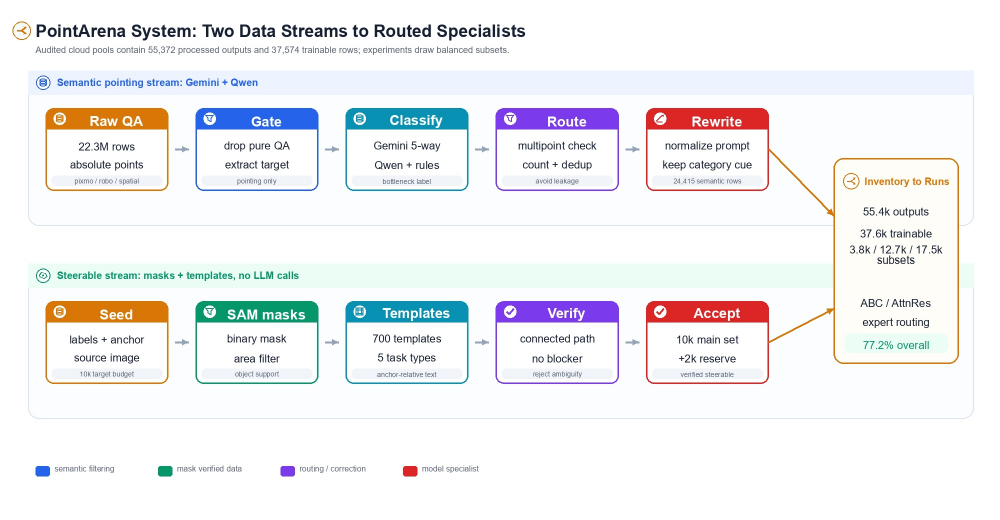}
  \caption{System overview. Two data streams create semantic and steerable supervision; ABC correction and AttnRes target different coordinate errors; category-aware routing selects the deployed expert per PointArena category.}
  \label{fig:overview}
\end{figure*}
\begin{itemize}
    \item \textbf{Agent-driven data synthesis.} Two data streams turn 22.3M raw QA rows into multi-batch cloud pools. The audited inventory scans 55,372 processed outputs and 53,772 unique IDs, yielding 37,574 trainable rows after latest-judgement de-duplication. Experiments then draw explicit balanced subsets: 3,815 rows for the early semantic baseline, 12,680 rows for mixed D-format runs, and up to 17,500 rows for steerability sweeps.
    \item \textbf{Targeted architecture and correction.} AttnRes adds gated cross-block attention for steerability. ABC correction injects perturbed points with visual features, improving coordinate grounding without relying on prompt-only metadata.
    \item \textbf{Category-aware routing.} Validation analysis shows that no checkpoint dominates all categories. The final submission routes Affordance/Counting/Spatial to the CoordMap expert, Reasoning to the PointMLP+ViT expert, and Steerability to AttnRes, reaching 77.2\% overall on the benchmark.
\end{itemize}

\begin{figure*}[t]
  \centering
  \includegraphics[width=.95\textwidth]{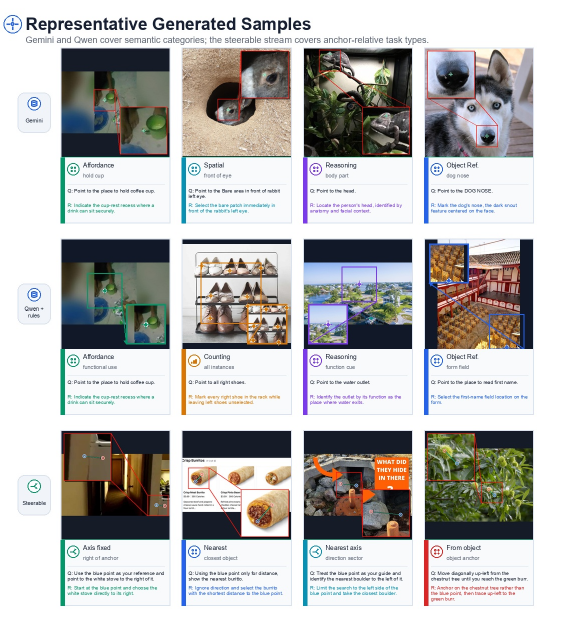}
  \caption{Representative generated samples with the original question and category-specific restatement. In each panel, \textbf{Q} is the original pointing question attached to the image, while \textbf{R} is our normalized restatement that makes the category cue explicit. Gemini and Qwen cover semantic categories; the steerable stream covers four anchor-relative task types. Full galleries are in \cref{app:samples}.}
  \label{fig:samples}
\end{figure*}

\cref{fig:samples} shows representative generated samples. The main paper focuses on the proof chain and the final numbers; full sample galleries, parameter sweeps, and diagnostic tables are in the appendix.

\section{Related Work}
\label{sec:related}

\textbf{Visual grounding and coordinate generation.}
Object detection models such as DETR~\cite{detr} localize boxes, while recent VLMs extend grounding to text-conditioned coordinates. Florence-2~\cite{florence2} unifies many vision-language tasks in promptable formats, and Molmo/PixMo~\cite{molmo_pixmo} provide open multimodal models and point-supervised data. PointArena~\cite{pointarena2026} differs from simple object pointing by separating affordance, spatial relation, reasoning, counting, and anchor-relative steerability.

\textbf{Synthetic supervision.}
Instruction tuning and VLM data synthesis are standard tools for broadening model behavior~\cite{llava,blip2,instructblip}. For pointing, however, the synthetic data must preserve valid coordinates and category-specific cognitive bottlenecks. Our data pipeline therefore combines LLM-based filtering and rewriting with rule-based routing, mask verification from Segment Anything-style models~\cite{sam}, and deterministic template generation.

\textbf{Spatial reasoning modules and refinement.}
Visual markers and point prompts can expose spatial structure to VLMs~\cite{som}. Iterative refinement is also common in detection, pose, and generative models~\cite{cascade-rcnn,simplebaseline,ddpm}. Our ABC variants test whether coordinate correction is helped by text-only metadata or by visually grounded point features; the experiments show that visual point encoding matters more than additional correction rounds. AttnRes is complementary: it changes cross-layer information flow rather than the coordinate input representation.

\section{Method}
\label{sec:method}

We fine-tune a Molmo-style 8B VLM~\cite{molmo_pixmo} with LoRA~\cite{lora}. The system has three moving parts: data synthesis, architecture changes for steerability, and point-conditioned correction.

\subsection{Agent-Driven Data Synthesis}
\label{sec:data-pipeline}

The data pipeline converts coordinate-bearing QA records into instructions aligned with PointArena categories. Each LLM-processed sample goes through four stages: gatekeeping, category classification, multipoint detection, and rewrite. The gatekeeper removes pure QA or non-pointing examples; classification assigns the cognitive bottleneck; multipoint detection routes all/every/several-style queries away from single-point data; rewrite normalizes the instruction while preserving the bottleneck.

\textbf{Semantic stream.} Gemini performs a broad five-class pass, while Qwen3-8B~\cite{qwen3} and rules add lower-cost local filtering, deduplication, and a counting track. \textbf{Steerable stream.} A deterministic generator filters labels, produces object masks with a SAM-style tracker~\cite{sam}, instantiates 700 anchor-relative templates, and rejects ambiguous paths with connected-component verification. The early semantic cache scans 37,498 Gemini-processed outputs and keeps 24,415 trainable completed rows. The broader steerable-D cache adds Qwen/nodup and verified SAM-clean steerable batches, scanning 55,372 outputs and retaining 37,574 trainable completed or accepted rows after de-duplication. We do not treat these as one monolithic training set: the 3,815-row value is only the early balanced semantic subset, while the later D-format and AttnRes experiments use 12,680-row mixed subsets and steerability sweeps up to 17,500 rows.

\subsection{AttnRes for Steerability}
\label{sec:attnres}

Steerability requires the model to retain an anchor state while following a direction. AttnRes adds a gated residual attention module every four transformer blocks. For hidden state $\mathbf{H}_i$ at block $i$,
\begin{equation}
    \mathbf{H}'_i = \mathbf{H}_i + \tanh(\alpha)
    \cdot \mathrm{AttnRes}(\mathbf{H}_i; \mathbf{H}_{i-3:i-1}),
    \label{eq:attnres}
\end{equation}
where the scalar gate $\alpha$ is initialized to zero. Thus training begins as the pretrained model and only opens the historical path if it helps. We use AttnRes as a specialist for Steerability rather than as a universal replacement, because the ablation shows that it can hurt non-steerable categories.

\subsection{ABC Correction}
\label{sec:abc}

ABC correction trains the model to recover a ground-truth point from a perturbed point. Given $\mathbf{p}^*$, we sample $\tilde{\mathbf{p}}=\mathbf{p}^*+\epsilon$ and ask the model to predict $\mathbf{p}^*$ conditioned on the image, instruction, and a representation of $\tilde{\mathbf{p}}$.

We compare three representations. \textbf{A} appends the perturbed coordinate as text. \textbf{B} encodes it with a PointMLP and fuses the result with intermediate ViT features. \textbf{C} adds a dense multi-scale coordinate heatmap processed by a small CNN. The key result is that visual point encoding, not the number of correction rounds, drives the gain: B and C peak at round 0, while extra rounds mostly improve stability rather than peak accuracy.

\SamePageWideFigure{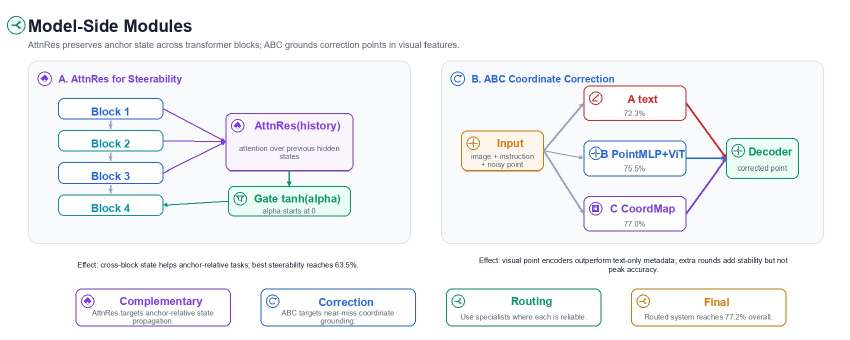}{Main model-side architecture. AttnRes adds a gated cross-block path for anchor-relative reasoning; ABC compares text-only point metadata with visually grounded point encoders.}{fig:model-arch}

\section{Experiments}
\label{sec:experiments}

\textbf{Setup.}
We fine-tune with LoRA rank 64, batch size 1, gradient accumulation 16, learning rate $10^{-4}$, and early stopping over 2k--20k checkpoints. Validation uses 982 PointArena samples across Affordance, Spatial Relation, Reasoning, Counting, and Steerability. A prediction is correct when the output point falls inside the target mask.

\begin{SamePageTable}{Main proof-chain results. Icons mark the intervention family; \isolatedbest{blue} marks the best isolated validation cell and \routedbest{green} marks the submitted routed system. For the routed row, category cells are local expert-selection validation scores, while All is the official submission score.}{tab:main-results}
  \scriptsize
  \setlength{\tabcolsep}{2.4pt}
  \begin{tabular}{@{}llcccccc@{}}
    \toprule
    Evidence & Config & Aff. & Cnt. & Rea. & Spa. & Ste. & All \\
    \midrule
    \evalicon & Zero-shot & 85.9 & \isolatedbest{73.0} & 77.2 & 76.9 & 50.5 & 72.7 \\
    \dataicon & Pipe-A & 93.9 & 67.3 & \isolatedbest{82.9} & 83.1 & 49.5 & 75.3 \\
    \abcicon & ABC-B & 93.9 & 68.9 & 79.8 & 79.0 & 61.5 & 75.5 \\
    \abcicon & ABC-C & \isolatedbest{94.4} & 69.9 & 79.3 & \isolatedbest{82.6} & 62.5 & 77.0 \\
    \routeicon & Routed$^\ast$ & \routedbest{93.9} & \routedbest{70.4} & \routedbest{78.2} & \routedbest{82.6} & \routedbest{63.0} & \routedbest{77.2} \\
    \bottomrule
  \end{tabular}
  \vspace{2pt}
  \parbox{.94\linewidth}{\scriptsize $^\ast$The routed category cells come from the validation evidence used to choose each expert in the submission report; the All cell reports the submitted benchmark score rather than a re-summed 982-sample validation ensemble.}
\end{SamePageTable}

\cref{tab:main-results} shows the main progression. Pipeline A proves that filtered semantic supervision is useful: it raises overall accuracy from 72.7\% to 75.3\% and gives large gains on Affordance, Reasoning, and Spatial Relation. The same row also exposes what the data alone does not fix: Counting falls and Steerability remains near 50\%. ABC-B/C then isolate the missing coordinate signal. Visual point encoding, especially CoordMap, recovers Counting and improves Steerability enough to reach 77.0\% overall. The final submission uses this analysis rather than a single blended checkpoint: CoordMap handles Affordance/Counting/Spatial, PointMLP+ViT handles Reasoning, and AttnRes handles Steerability.

\textbf{Attribution.}
AttnRes is a targeted steerability fix: at steerable weight $n{=}2$, it improves Steerability from 59.5\% to 63.5\%, but the gain disappears at $n{=}1$ and $n{=}4$. Point encoding is the main correction signal: text-only correction gives 72.3\%, PointMLP+ViT gives 75.5\%, and CoordMap reaches 77.0\%. Extra correction rounds do not improve the best checkpoint, so the useful signal is visual grounding of the perturbed point rather than repeated inference. The appendix separates final-supporting evidence from diagnostics: early semantic-only runs, weak or collapsed checkpoints, prompt-only controls, and SinglePOS variants are included to explain rejected routes rather than to claim additional submission systems.

\section{Conclusion}
\label{sec:conclusion}

We presented a compact PointArena 2026 system built around targeted supervision and category-specific experts. The data pipeline supplies semantic diversity and a new source of steerable examples; AttnRes improves anchor-relative reasoning when paired with the right steerable data scale; ABC correction shows that visually grounded point encoding is more important than repeated correction rounds. The final routed system reaches 77.2\% overall and ranks second. The largest remaining gaps are Counting, where errors are mostly wrong cardinality or out-of-mask points, and Steerability, where cross-block state propagation still leaves substantial headroom.

\clearpage
\appendix
\section{Qualitative Galleries}
\label{app:samples}

\SamePageWideFigure[.9\textwidth]{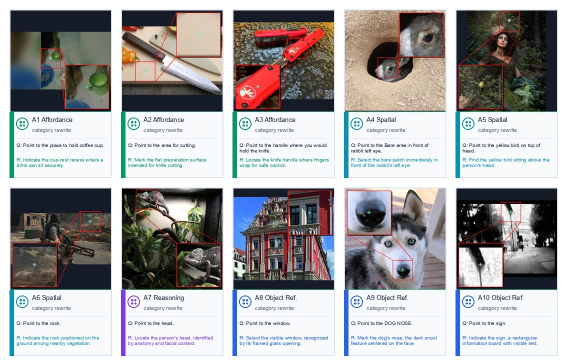}{Gemini balanced examples. Each panel preserves the full scene, adds a local zoom around the target point, and shows \textbf{Q} as the original question and \textbf{R} as the category-specific restatement used to expose the intended bottleneck.}{fig:app-pipeline-a}

\clearpage
\SamePageWideFigure[.9\textwidth]{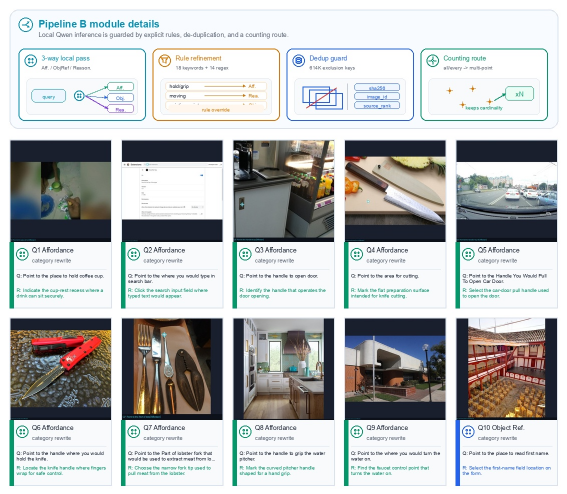}{Qwen/rule examples with module-level detail. The top row visualizes the local classifier, rule refinement, de-duplication guard, and counting route; in the sample grid, \textbf{Q} is the raw input question and \textbf{R} is the rule/category-aware restatement.}{fig:app-qwen}

\clearpage
\SamePageWideFigure[.9\textwidth]{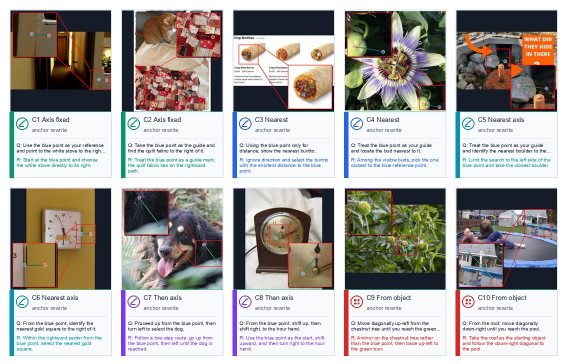}{Pipeline C steerable examples. The blue point is the anchor and the red point is the target; \textbf{Q} is the synthesized anchor-relative question, while \textbf{R} restates the same task in the corresponding steerable subtype language.}{fig:app-steerable}

\clearpage
\section{Architecture Diagrams}
\label{app:architecture}

\SamePageWideFigure{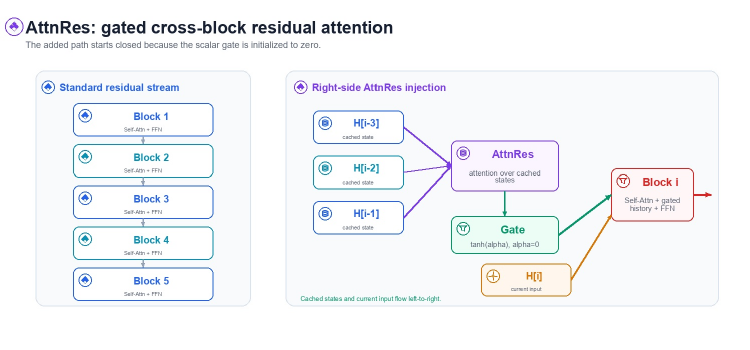}{Detailed AttnRes architecture. The gate is initialized at zero so the added cross-block path starts as an identity residual.}{fig:app-attnres}

\clearpage
\SamePageWideFigure{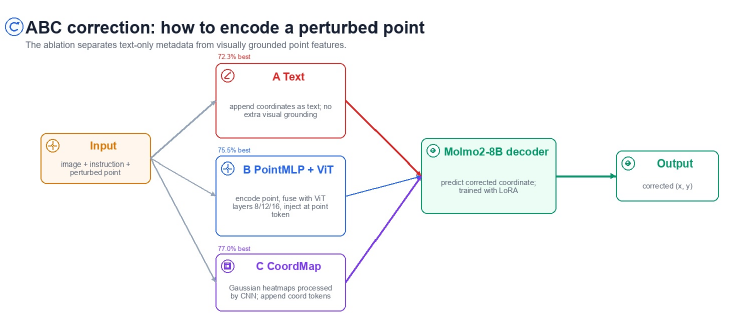}{Detailed ABC coordinate encoding. B grounds a perturbed point through PointMLP and visual features; C adds a dense coordinate map.}{fig:app-abc}

\clearpage
\section{Detailed Experimental Tables}
\label{app:experiments}

The additional tables report the negative results as well as the positive ones. They show three recurring patterns: early checkpoints are usually best, coordinate likelihood can improve while mask accuracy declines, and format-level shortcuts rarely help unless the coordinate representation is visually grounded. Icons group the evidence: \dataicon{} data, \evalicon{} optimization/evaluation, \attnicon{} AttnRes, \abcicon{} ABC, \ctrlcon{} negative controls, and \routeicon{} final routing implications.

\subsection{Data Funnel}

\noindent\textbf{What these tables contain.}
This group is bookkeeping rather than model ranking. It separates raw cloud outputs, latest trainable rows, and the balanced subsets actually sampled by each experiment. The important distinction is that 3,815 rows are only the early semantic-only LoRA subset; later D-format, ABC, SinglePOS, and AttnRes studies draw from the larger audited pool.

\begin{SamePageTable}{\dataicon{} Audited data accounting. The 3,815-row value is a small balanced subset, not the server-side sample count.}{}
  \centering\scriptsize
  \setlength{\tabcolsep}{2pt}
  \begin{tabular}{@{}llrrl@{}}
    \toprule
    & Pool & Scanned & Trainable & Note \\
    \midrule
    \dataicon & Gemini cache & 37,498 & 24,415 & early LLM pool \\
    \dataicon & \textbf{D cache} & \bestcell{55,372} & \bestcell{37,574} & Gemini+Qwen+SAM \\
    \dataicon & Unique IDs in D cache & \gaincell{53,772} & -- & after sample-id merge \\
    \evalicon & Early semantic train subset & -- & \riskcell{3,815} & 5 classes, 763 each \\
    \evalicon & Mixed D train subset & -- & 12,680 & 5 classes, 2,536 each \\
    \attnicon & Largest sweep subset & -- & 17,500 & 10k steer + semantic mix \\
    \bottomrule
  \end{tabular}
\end{SamePageTable}

\begin{SamePageTable}{\dataicon{} Trainable rows in the merged steerable-D cache. This is the better answer to ``how many usable samples are on the server'' after latest-judgement de-duplication.}{}
  \centering\footnotesize
  \setlength{\tabcolsep}{5pt}
  \begin{tabular}{@{}llr@{}}
    \toprule
    & Category & Trainable rows \\
    \midrule
    \dataicon & Counting & \bestcell{13,663} \\
    \attnicon & Steerable & \bestcell{10,000} \\
    \dataicon & Object Reference & 7,307 \\
    \dataicon & Affordance & 3,346 \\
    \dataicon & Reasoning & 3,258 \\
    \evalicon & Spatial Relation & \riskcell{$0^{\dagger}$} \\
    \midrule
    & Total & \bestcell{37,574} \\
    \bottomrule
  \end{tabular}
  \par\vspace{2pt}
  \raggedright\scriptsize $^{\dagger}$The active D-cache policy filters completed Spatial Relation rows out of this merged set.
\end{SamePageTable}

\begin{SamePageTable}{\dataicon{} Why the early semantic baseline is only 3,815 rows. Balancing capped all classes by the rarest eligible category, Spatial Relation.}{}
  \centering\footnotesize
  \setlength{\tabcolsep}{5pt}
  \begin{tabular}{@{}llrrr@{}}
    \toprule
    & Category & Eligible & Selected & Discarded \\
    \midrule
    \dataicon & Affordance & 1,778 & 763 & 1,015 \\
    \dataicon & Counting & \bestcell{12,895} & 763 & \riskcell{12,132} \\
    \dataicon & Object Reference & 4,224 & 763 & 3,461 \\
    \dataicon & Reasoning & 2,478 & 763 & 1,715 \\
    \dataicon & Spatial Relation & \riskcell{763} & \riskcell{763} & 0 \\
    \bottomrule
  \end{tabular}
\end{SamePageTable}

\begin{SamePageTable}{\dataicon{} Server-side semantic/local batch inventory. These are processed Gemini and Qwen/rule outputs before latest-result selection; they feed semantic categories and should not be summed as final train rows.}{}
  \centering\scriptsize
  \setlength{\tabcolsep}{1.7pt}
  \begin{tabular}{@{}llrl@{}}
    \toprule
    & Batch & Output & Accepted/latest role \\
    \midrule
    \dataicon & Gemini non-count & 10,005 & 6,918 D / 7,444 early \\
    \dataicon & Gemini pointsabs & 6,329 & 2,594 D / 2,872 early \\
    \dataicon & Gemini rulefilter & \bestcell{12,003} & 6,757 rerun pool \\
    \dataicon & Gemini robopoint & 3,204 & 3,201 candidates \\
    \dataicon & Qwen nodup & 4,879 & 1,594 local dedup \\
    \dataicon & Qwen add2000 & 3,005 & 2,581 object-ref add-on \\
    \dataicon & Qwen 20k v2 & 2,005 & 1,072 Aff./Reason. add-on \\
    \bottomrule
  \end{tabular}
\end{SamePageTable}

\begin{SamePageTable}{\dataicon{} Server-side Counting and Steerable batch inventory. Counting rows are candidate scans later sampled by training scripts; SAM-clean rows are verified anchor-relative examples used in steerability studies.}{}
  \centering\scriptsize
  \setlength{\tabcolsep}{2pt}
  \begin{tabular}{@{}llrl@{}}
    \toprule
    & Batch & Output & Accepted/latest role \\
    \midrule
    \dataicon & Gemini counting cont. & 4,196 & 3,085 count candidates \\
    \dataicon & Counting direct v1--v4 & \bestcell{50,000} & scans, later sampled \\
    \attnicon & SAM-clean main & \bestcell{10,000} & \bestcell{verified steerable} \\
    \attnicon & SAM-clean reserve & 2,000 & held-out reserve \\
    \bottomrule
  \end{tabular}
\end{SamePageTable}

\begin{SamePageTable}{\dataicon{} Training subset sizes used by experiment family. Experiments resample balanced subsets from the larger audited pools, so subset rows should not be read as total generated samples.}{}
  \centering\scriptsize
  \setlength{\tabcolsep}{2.5pt}
  \begin{tabular}{@{}llrl@{}}
    \toprule
    & Family & Rows & Composition \\
    \midrule
    \evalicon & Early filtered & \riskcell{3,815} & 5 semantic classes, 763 each \\
    \evalicon & Archived filtered & 4,865 & 5 semantic classes, 973 each \\
    \abcicon & Mixed D / ABC / SinglePOS & 12,680 & 5 classes, 2,536 each \\
    \attnicon & Steer sweep $n{=}1$ & 10,144 & 4 classes, 2,536 each \\
    \attnicon & Steer sweep $n{=}2$ & 12,680 & steerable doubled to 5,072 \\
    \attnicon & Steer sweep $n{=}4$ & \bestcell{17,500} & 10k steerable + 3 semantic classes \\
    \bottomrule
  \end{tabular}
\end{SamePageTable}

\subsection{Optimization and Training Dynamics}

\noindent\textbf{What these tables contain.}
These are early filtered-data fine-tuning runs on the 3,815-row semantic subset. They do not include a dedicated steerable training class, so Steerability is reported only as an evaluation stress test. The weak and collapsed rows are kept because they justify the chosen optimizer regime and show why semantic filtering alone cannot solve anchor-relative pointing.

\begin{SamePageTable}{\evalicon{} Filtered-data fine-tuning grid on the 3,815-row semantic subset. LR 2e-4 with GA 8 collapses the output format; larger effective batch is stable, but Steerability remains near chance because this subset has no steerable supervision.}{}
  \centering\footnotesize
  \setlength{\tabcolsep}{3pt}
  \begin{tabular}{@{}llcccccc@{}}
    \toprule
    & Run & Overall & Aff. & Spatial & Reason. & Steer. & Count. \\
    \midrule
    \evalicon & lr1e-4, GA16 & \bestcell{75.25} & \bestcell{93.94} & \bestcell{83.08} & \bestcell{82.90} & 49.5 & 67.35 \\
    \evalicon & lr2e-4, GA16 & 74.85 & 90.91 & 81.54 & 78.24 & \bestcell{54.0} & \bestcell{69.90} \\
    \evalicon & lr1e-4, GA8 & 73.83 & 93.43 & \bestcell{83.08} & 76.17 & 52.0 & 64.80 \\
    \ctrlcon & lr2e-4, GA8 & \riskcell{14.87} & \riskcell{31.82} & \riskcell{16.92} & \riskcell{10.88} & \riskcell{12.5} & \riskcell{2.04} \\
    \bottomrule
  \end{tabular}
\end{SamePageTable}

\begin{SamePageTable}{\evalicon{} Training trajectories for the two stable semantic-only runs. Accuracy peaks early while train loss continues to fall, indicating overfitting to the synthetic distribution rather than better PointArena mask hits.}{}
  \centering\scriptsize
  \setlength{\tabcolsep}{3pt}
  \begin{tabular}{@{}llrrrrrrr@{}}
    \toprule
    & Run/step & Overall & Aff. & Spatial & Reason. & Steer. & Count. & Loss \\
    \midrule
    \evalicon & lr1e-4 GA16 2k & \bestcell{75.25} & 93.94 & \bestcell{83.08} & \bestcell{82.90} & 49.5 & 67.35 & 0.553 \\
    \evalicon & lr1e-4 GA16 4k & 72.40 & \bestcell{94.95} & 68.72 & 77.72 & 53.5 & 67.35 & 0.482 \\
    \evalicon & lr1e-4 GA16 8k & 70.57 & 92.93 & 63.59 & 80.83 & 53.0 & 62.76 & 0.434 \\
    \evalicon & lr1e-4 GA16 16k & 69.65 & 92.93 & 65.64 & 78.76 & 51.0 & 60.20 & \riskcell{0.311} \\
    \evalicon & lr2e-4 GA16 2k & 74.85 & 90.91 & 81.54 & 78.24 & 54.0 & \bestcell{69.90} & 0.540 \\
    \evalicon & lr2e-4 GA16 4k & 73.83 & 92.42 & 77.95 & 78.24 & \bestcell{55.5} & 65.31 & 0.487 \\
    \evalicon & lr2e-4 GA16 8k & 72.10 & 91.92 & 76.41 & 77.72 & 51.0 & 63.78 & 0.441 \\
    \evalicon & lr2e-4 GA16 16k & 70.47 & 90.40 & 76.41 & 76.17 & 48.0 & 61.73 & \riskcell{0.311} \\
    \bottomrule
  \end{tabular}
\end{SamePageTable}

\subsection{Module Ablations}

\noindent\textbf{What these tables contain.}
These ablations isolate mechanisms, not full submissions. The AttnRes rows vary the amount of verified steerable data; the ABC rows vary how perturbed points are visually encoded; the large D-format and loss tables show why a single mixed model is weaker than routing complementary specialists.

\begin{SamePageTable}{\attnicon{} AttnRes steerable-data sweep. Each row compares the default backbone and the cross-block AttnRes variant under the same steerable-data weight, isolating when the module helps.}{}
  \centering\scriptsize
  \setlength{\tabcolsep}{2.5pt}
  \begin{tabular}{@{}llcccl@{}}
    \toprule
    & Weight & Default & Step & AttnRes & Interpretation \\
    \midrule
    \attnicon & $n{=}1$ & 62.5 & 2k & \riskcell{60.0} & too little steerable signal \\
    \attnicon & $n{=}2$ & 59.5 & 10k & \bestcell{63.5} & best cross-block regime \\
    \attnicon & $n{=}4$ & 60.0 & 14k & \riskcell{58.0} & repeated data overfits \\
    \bottomrule
  \end{tabular}
\end{SamePageTable}

\begin{SamePageTable}{\attnicon{} AttnRes plus point-encoding controls. Adding anchor PointMLP+ViT does not beat pure AttnRes, so the final steerability expert keeps the simpler cross-block module.}{}
  \centering\scriptsize
  \setlength{\tabcolsep}{3pt}
  \begin{tabular}{@{}llccc@{}}
    \toprule
    & Variant & Extra encoding & Correction & Best Steer. \\
    \midrule
    \attnicon & AttnRes $n{=}2$ & none & no & \bestcell{63.5} \\
    \abcicon & AttnRes-B & PointMLP+ViT & no & 62.0 \\
    \abcicon & AttnRes-B-iter & PointMLP+ViT & yes & 62.0 \\
    \bottomrule
  \end{tabular}
\end{SamePageTable}

\begin{SamePageTable}{\abcicon{} ABC correction rounds on mixed D-format data. Iteration is not the main source of improvement; point-conditioned visual grounding is.}{}
  \centering\footnotesize
  \setlength{\tabcolsep}{5pt}
  \begin{tabular}{@{}llcccc@{}}
    \toprule
    & Variant & R0 & R1 & R2 & Best round \\
    \midrule
    \abcicon & A text-only & 72.3 & 72.2 & 72.0 & R0 \\
    \abcicon & B PointMLP+ViT & 75.5 & 75.5 & 75.5 & tie \\
    \abcicon & C +CoordMap & \bestcell{77.0} & 75.5 & 76.7 & R0 \\
    \bottomrule
  \end{tabular}
\end{SamePageTable}

\begin{SamePageTable}{\abcicon{} ABC best category scores. CoordMap mainly helps Spatial; PointMLP+ViT is competitive for Reasoning with fewer parameters, which motivates category-aware expert selection.}{}
  \centering\scriptsize
  \setlength{\tabcolsep}{2pt}
  \begin{tabular}{@{}llrrrrrr@{}}
    \toprule
    & Variant & Overall & Aff. & Count. & Reason. & Spatial & Steer. \\
    \midrule
    \abcicon & A text-only & 72.3 & 90.9 & 61.7 & 72.5 & 80.0 & 61.5 \\
    \abcicon & B PointMLP+ViT & 75.5 & 93.9 & 68.9 & \bestcell{79.8} & 79.0 & 61.5 \\
    \abcicon & C CoordMap & \bestcell{77.0} & \bestcell{94.4} & \bestcell{69.9} & 79.3 & \bestcell{82.6} & \bestcell{62.5} \\
    \bottomrule
  \end{tabular}
\end{SamePageTable}

\begin{SamePageTable}{\routeicon{} Large D-format mixed training. This 12,680-row run uses 2,536 examples per class, including steerable data; it improves Steerability relative to Pipeline A but loses overall accuracy and Counting, motivating expert routing rather than a single enlarged mixed model.}{}
  \centering\scriptsize
  \setlength{\tabcolsep}{3pt}
  \begin{tabular}{@{}llrrrrrrr@{}}
    \toprule
    & Step & Overall & Aff. & Spatial & Reason. & Steer. & Count. & Train loss \\
    \midrule
    \routeicon & 2k & \bestcell{73.42} & \bestcell{93.43} & 75.90 & 75.13 & 55.50 & \bestcell{67.35} & 0.5666 \\
    \routeicon & 14k & 73.32 & 89.39 & \bestcell{81.03} & 74.09 & \bestcell{59.50} & 62.76 & 0.4687 \\
    \routeicon & 16k & 72.91 & 90.91 & 76.41 & 77.20 & 57.00 & 63.27 & 0.4574 \\
    \routeicon & 20k & 73.01 & 91.41 & 77.44 & \bestcell{77.72} & 57.00 & \riskcell{61.73} & \riskcell{0.4527} \\
    \bottomrule
  \end{tabular}
\end{SamePageTable}

\begin{SamePageTable}{\evalicon{} Loss diagnostics for the mixed D lr2e-4 GA16 run. Coordinate CE decreases, but PointArena accuracy does not improve; late training mostly optimizes token likelihood rather than mask correctness.}{}
  \centering\scriptsize
  \setlength{\tabcolsep}{2pt}
  \begin{tabular}{@{}llrrrrr@{}}
    \toprule
    & Ckpt. & Train & Coord & Label & All & Count. \\
    \midrule
    \evalicon & 2k & 0.5666 & 1.4361 & 0.5425 & \bestcell{73.42} & \bestcell{67.35} \\
    \evalicon & 14k & 0.4687 & 1.3498 & 0.2538 & 73.32 & 62.76 \\
    \evalicon & 16k & 0.4574 & 1.3308 & 0.2041 & 72.91 & 63.27 \\
    \evalicon & 20k & \bestcell{0.4527} & \bestcell{1.3190} & \bestcell{0.1757} & 73.01 & \riskcell{61.73} \\
    \bottomrule
  \end{tabular}
\end{SamePageTable}

\begin{SamePageTable}{\evalicon{} Coordinate digit loss at lr2e-4 GA16. The ones digit remains the hardest coordinate component, explaining why lower CE does not guarantee mask hits.}{}
  \centering\footnotesize
  \setlength{\tabcolsep}{5pt}
  \begin{tabular}{@{}llrrrr@{}}
    \toprule
    & Checkpoint & Coord all & Hundreds & Tens & Ones \\
    \midrule
    \evalicon & 2k & 1.4361 & 0.7050 & 1.4679 & \riskcell{2.1359} \\
    \evalicon & 14k & 1.3498 & 0.5921 & 1.3918 & \riskcell{2.0661} \\
    \evalicon & 16k & 1.3308 & 0.5816 & 1.3758 & \riskcell{2.0356} \\
    \evalicon & 20k & \bestcell{1.3190} & \bestcell{0.5680} & \bestcell{1.3600} & \riskcell{2.0295} \\
    \bottomrule
  \end{tabular}
\end{SamePageTable}

\subsection{Negative Controls and Error Anatomy}

\noindent\textbf{What these tables contain.}
These rows are negative controls and failure analysis. Prompt-only metadata variants test whether coordinate hints alone help; SinglePOS replaces Molmo2's native continuous HTML coordinate protocol with discrete POS tokens; the Counting table explains the dominant remaining error modes. None of these rows is a final candidate system.

\begin{SamePageTable}{\ctrlcon{} Prompt-only coordinate metadata controls. These runs change the prompt-level coordinate metadata without changing the model architecture, showing that formatting alone does not provide reliable grounding.}{}
  \centering\scriptsize
  \setlength{\tabcolsep}{1.5pt}
  \begin{tabular}{@{}llccc@{}}
    \toprule
    & Prompt & Overall & Steer. & Conclusion \\
    \midrule
    \ctrlcon & NL coords & 69.55 & 40.5 & hurts steering \\
    \ctrlcon & NL+HTML & 69.04 & 38.0 & format mismatch \\
    \evalicon & No injection & 71.38 & 49.5 & baseline prompt \\
    \ctrlcon & Legacy list & \bestcell{71.69} & \bestcell{50.5} & best prompt \\
    \ctrlcon & Bare HTML tag & \riskcell{68.33} & \riskcell{34.5} & disrupts output \\
    \ctrlcon & Normalized list & 70.57 & 45.5 & below baseline \\
    \bottomrule
  \end{tabular}
\end{SamePageTable}

\begin{SamePageTable}{\ctrlcon{} SinglePOS coordinate-token controls on the 12,680-row mixed D set. Discrete coordinate tokens underperform Molmo2's continuous HTML coordinate protocol and several variants collapse, so SinglePOS is a rejected encoding route.}{}
  \centering\footnotesize
  \setlength{\tabcolsep}{3pt}
  \begin{tabular}{@{}llcccl@{}}
    \toprule
    & Variant & Overall & Spatial & Count. & Outcome \\
    \midrule
    \evalicon & HTML baseline & \bestcell{73.42} & \bestcell{75.9} & \bestcell{67.3} & stable \\
    \ctrlcon & SinglePOS original & 64.56 & 69.7 & 40.8 & still rising \\
    \ctrlcon & SinglePOS no-count & \riskcell{13.95} & \riskcell{16.4} & \riskcell{1.5} & no count prior \\
    \ctrlcon & Legacy no-count & 56.11 & 63.6 & 32.1 & epoch-2 collapse \\
    \ctrlcon & Count + legacy & \riskcell{14.66} & \riskcell{13.8} & \riskcell{2.6} & token dilution \\
    \bottomrule
  \end{tabular}
\end{SamePageTable}

\begin{SamePageTable}{\evalicon{} Counting error anatomy at the best early checkpoint. This table decomposes the 196 Counting validation samples rather than aggregating across checkpoints.}{}
  \centering\footnotesize
  \setlength{\tabcolsep}{5pt}
  \begin{tabular}{@{}llc@{}}
    \toprule
    & Metric & Value \\
    \midrule
    \evalicon & Counting samples & 196 \\
    \evalicon & Success & \bestcell{132 / 196 = 67.35\%} \\
    \ctrlcon & Wrong number of points & \riskcell{50 / 196 = 25.51\%} \\
    \ctrlcon & At least one point out of mask & \riskcell{44 / 196 = 22.45\%} \\
    \ctrlcon & Correct count but out of mask & 14 / 146 = 9.59\% \\
    \ctrlcon & Wrong count and out of mask & \riskcell{30 / 50 = 60.00\%} \\
    \ctrlcon & Invalid parse & 5 / 196 = 2.55\% \\
    \bottomrule
  \end{tabular}
\end{SamePageTable}

\end{document}